\documentclass[letterpaper]{article} 
\usepackage{aaai25}  
\usepackage{times}  
\usepackage{helvet}  
\usepackage{courier}  
\usepackage[hyphens]{url}  
\usepackage{graphicx} 
\usepackage{natbib}  
\usepackage{caption} 
\usepackage{algorithm}
\usepackage{algorithmic}
\usepackage{tabularray}
\usepackage{amsmath}
\UseTblrLibrary{siunitx}

\usepackage{newfloat}
\usepackage{listings}
\DeclareCaptionStyle{ruled}{labelfont=normalfont,labelsep=colon,strut=off} 
\floatstyle{ruled}
\newfloat{listing}{tb}{lst}{}
\floatname{listing}{Listing}
\title{Tradutor: Building a Variety Specific Translation Model}
\author {
    Hugo Sousa\textsuperscript{\rm 1,\rm 2}\equalcontrib,
    Satya Almasian\textsuperscript{\rm 3}\equalcontrib,
    Ricardo Campos\textsuperscript{\rm 2,\rm 4,\rm 5},
    Alípio Jorge\textsuperscript{\rm 1,\rm 2}
}
\affiliations {
    \textsuperscript{\rm 1}Faculty of Sciences, University of Porto, Porto, Portugal\\
    \textsuperscript{\rm 2}INESC TEC, Porto, Portugal\\
    \textsuperscript{\rm 3}Institute of Computer Science, Heidelberg University, Germany\\
    \textsuperscript{\rm 4}Department of Informatics, University of Beira Interior, Covilhã, Portugal \\
    \textsuperscript{\rm 5}Ci2 - Smart Cities Research Center, Tomar, Portugal \\
    hugo.o.sousa@inesctec.pt, almasian@informatik.uni-heidelberg.de
}

\begin{document}

\maketitle

\begin{abstract}
Language models have become foundational to many widely used systems. However, these seemingly advantageous models are double-edged swords. While they excel in tasks related to resource-rich languages like English, they often lose the fine nuances of language forms, dialects, and varieties that are inherent to languages spoken in multiple regions of the world. Languages like European Portuguese are neglected in favor of their more popular counterpart, Brazilian Portuguese, leading to suboptimal performance in various linguistic tasks. To address this gap, we introduce the first open-source translation model specifically tailored for European Portuguese, along with a novel dataset specifically designed for this task. Results from automatic evaluations on two benchmark datasets demonstrate that our best model surpasses existing open-source translation systems for Portuguese and approaches the performance of industry-leading closed-source systems for European Portuguese. By making our dataset, models, and code publicly available, we aim to support and encourage further research, fostering advancements in the representation of underrepresented language varieties.
\end{abstract}

\section{Introduction}

In an era, where Language Models (LMs) form the foundation of numerous tools and systems, a significant concern arises: how can we ensure these tools are equally beneficial to all communities? One major hurdle in this direction is \emph{inclusiveness}, particularly when it comes to scarce language varieties and the lack of models that take into account the diversity of dialects and culturally significant nuances of a given language~\cite{CostaJussa2018}. 
For example, while Portuguese is not considered a low-resource language, most of the content available is in Brazilian Portuguese. 
This is due to the Brazilian population representing more than 70\% of all native Portuguese speakers\footnote{Statistic derived from Wikipedia \url{https://en.wikipedia.org/wiki/Portuguese_language#Lusophone_countries}}. 
Consequently, a LM or a translation system trained on a Portuguese corpus is inclined to produce text containing phrases and grammatical structures of Brazilian Portuguese.  
As a result, countries like Portugal and Mozambique, which use different varieties of Portuguese, could find themselves at a disadvantage, particularly in deploying LM-based systems in critical areas such as healthcare and judiciary, where the grammar and lexicons of the language are of great importance~\cite{Scherre2016, Kato2016, Brito2016}.

One solution would be to create LMs specific to a certain language variety. However, this presents its own set of challenges for training and evaluation~\cite{ArmengolEstape2021,Rodrigues2023}. 
Training a LM, be it from scratch or fine-tuning,  requires a large, carefully curated corpus for training and several benchmarks for evaluation~\cite{Albalak2024}, which are either nonexistent or contaminated by the dominant language varieties. 
Another way to overcome these challenges is to create machine translation (MT) models dedicated to a specific language variety~\cite{Zbib2012, Sennrich2016, Riley2023}. 
With a sufficiently powerful MT model for the low-resource variety, we can take the first step toward resource creation and inclusion of these languages.
Such models can be used to translate training and evaluation benchmarks, which are predominantly in English, to the low-resource variety. 
Additionally, they can serve as an off-the-shelf intermediary between widely-used LMs and systems in low-resource settings or even be used to artificially generate data in the desired language variety.

In this paper, we present a novel methodology for developing a neural machine translation (NMT) model tailored to low-resource language varieties, where manually annotated data is scarce or unavailable. Our approach leverages a retro-translation technique: we first gather texts in the low-resource language variety and translate them into a resource-rich language. This newly created parallel corpus is then used to fine-tune a pre-trained language model. We validate our methodology using European Portuguese as a case study, though the same steps can be applied to other languages with similar challenges. As part of this work, we have created and publicly released a meticulously curated parallel corpus for European Portuguese, comprising 1,719,002 documents -- the largest of its kind to date. We evaluate our model against existing open and close source translation systems for Portuguese, as well as zero-shot language models, demonstrating the effectiveness of our approach. Finally, to ensure our translations remain true to the intended language variety, we use a language variety classification model to quantify if the texts produced by our translation model are effectively European Portuguese. 

In summary, our contributions are the following:

\begin{enumerate}

  \item We propose a methodology to create a parallel corpus for a low-resource language variety using an on-the-shelf translation model. 
  To this end, we provide the community with the largest translation dataset for European Portuguese and English, named PTradutor.
  
  \item We provide the first  open-source translation models from English to European Portuguese which  outperform the generic Portuguese open-source translation systems and close the gap to state-of-the-art close-source translations systems for European Portuguese.
  
  \item We offer a comprehensive evaluation of our models, emphasizing not only translation quality but also linguistic alignment to the desired language variety.

\end{enumerate}

The remainder of this manuscript is organized as follows: Section~\ref{sec:ptradutor} offers a comprehensive overview of the development of our corpus, PTradutor, which serves as the foundation for training our models. Sections~\ref{sec:exp_setup} and \ref{sec:eval} describe the experimental setup and present the results of our experiments, respectively. In Section~\ref{sec:related_work}, we position our research within the context of existing approaches. Finally, Section~\ref{sec:conculsion} summarizes our findings and suggests directions for future research.

\section{PTradutor}
\label{sec:ptradutor}

As noted previously, a significant challenge in training a translation model for a specific language variety is the scarcity of datasets tailored for this purpose. To overcome this obstacle, we devised a three-step approach to automatically generate a parallel corpus. Industry-grade translation systems typically achieve higher translation quality when translating \emph{into} a resource-rich language, such as English, rather than a low-resource language, such as European Portuguese. We exploit this concept to transform a monolingual European Portuguese corpus into a high-quality parallel corpus, using the following method:

\begin{enumerate}
    
    \item \textbf{Mono-lingual Corpus:} We compiled a large mono-lingual collection of texts written in European Portuguese.
    
    \item \textbf{Translation:} Using a translation system, we created a parallel corpus from the mono-lingual corpus, by translating from European Portuguese to English.
    
    \item \textbf{Filtering:} We conducted filtering and quality checks to ensure the integrity of the dataset.
\end{enumerate}

This process creates a corpus with parallel data pairs for English and European Portuguese, enabling the training of MT systems. In the following sections, we describe each step in detail.

\subsection{European Portuguese Corpus Collection}
For the collection of texts in European Portuguese, we used two sources: the DSL-TL~\cite{Zampieri2024} and the PtBrVid corpus~\cite{Sousa2025}. 
The DSL-TL corpus comprises a total of 4,458 news articles\footnote{Although the authors report 4,953 articles in their paper, the files shared on the project repository \url{https://github.com/LanguageTechnologyLab/DSL-TL} contain 4,458 rows.} written in Portuguese, manually annotated as ``European Portuguese'', ``Brazilian Portuguese'', or ``Both''. 
For our purposes, we use the train partition of this dataset and keep the texts labeled as ``European Portuguese'' and ``Both'' as they are both valid texts in European Portuguese, resulting in a total of 1,734 documents.

The larger chunk of our data comes from the PtBrVid corpus, which was originally created to train a Portuguese variety identifier that discriminates between European and Brazilian Portuguese. 
This corpus was constructed by compiling several datasets for each variety across six domains -- journalism, web, social media, literature, legal, and politics -- using metadata from the original datasets to label the variety of its texts. 
For example, the political subset uses the CETEM Público~\cite{Rocha2000} corpus for European Portuguese and the CETEM Folha\footnote{\url{https://linguateca.pt/cetenfolha/index_info.html}} for Brazilian Portuguese. 
For this research, we kept all entries that were labeled as ``European Portuguese''.

\subsection{Translation}

Monolingual data was translated into English using Google Translate with the Python library \texttt{ deep\_translator}\footnote{\url{https://github.com/nidhaloff/deep-translator}}. The reason for this choice is the accessibility and relatively good quality of Google Translate, however, this step can be achieved by other translation systems. When this dataset was created, Google Translate did not distinguish between European and Brazilian Portuguese. Consequently, using it as the translation engine assumes that the translation from European Portuguese to English is unaffected by this limitation. 

To verify this hypothesis, we conducted an experiment using one of our test datasets, namely FRMT~\cite{Riley2023}, which contains 2,616 examples in English, Brazilian Portuguese, and European Portuguese.  We used Google Translate to translate the European Portuguese and Brazilian Portuguese texts into English and then compared the English translations produced from the different varieties.  We found that Google Translate produced exactly the same translation for approximately $87.2\%$ of the dataset and achieved a BLEU score of $96.8\%$.  This confirms that the quality of the dataset is only mildly affected by Google’s lack of differentiation between varieties. Nonetheless, this assumption is language-specific and might not hold for different language varieties and different translation systems.

\subsection{Filtering}
The aggregation of the mentioned resources results in a total of $3,966,538$ documents. 
As previous research focused on training and fine-tuning LMs has emphasized the importance of data quality~\cite{Wenzek2020,Penedo2024}, we apply extensive filtering and quality checks to our dataset. 

\begin{figure}[htbp]
    \centering
    \includegraphics[width=\linewidth]{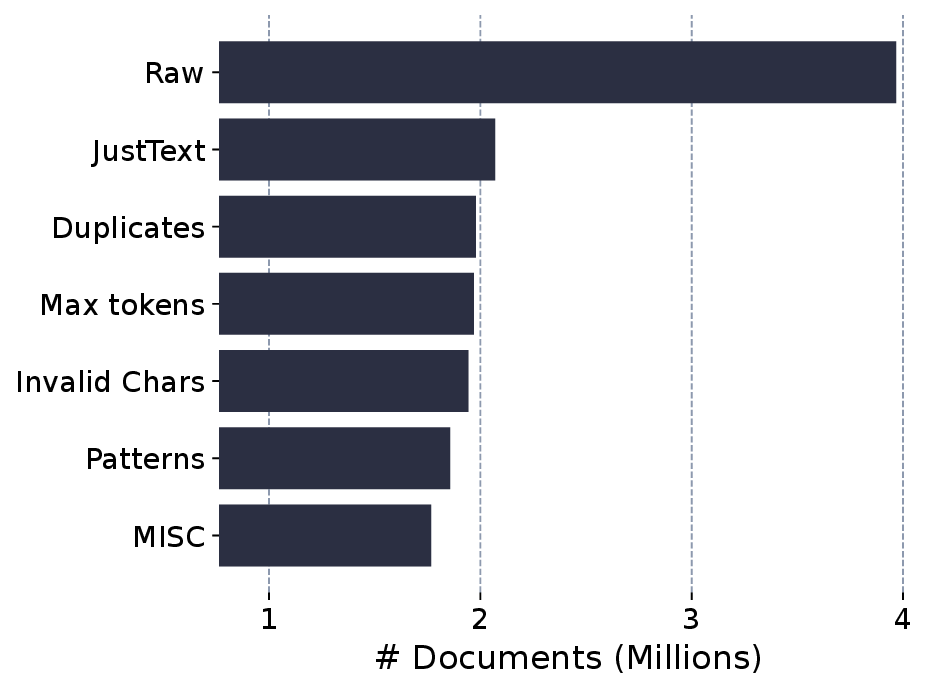}
    \caption{Number of documents (in millions) remaining after each step of our filtering pipeline.}
    \label{fig:filter_pipeline}
\end{figure}

Specifically, our filtering pipeline begins by using the \texttt{jusText}\footnote{\url{https://github.com/miso-belica/jusText}} library -- designed to clean boilerplate content by classifying textual blocks based on various features such as length and stop word density --  to remove entries classified as low-quality text. As depicted in Figure~\ref{fig:filter_pipeline}, this is the most stringent filter in our process, eliminating approximately 1.9 million documents from the collection. A closer examination reveals that most of the discarded documents originated from the Social Media partition of the PtBrVId corpus, which initially contained 2,014,752 documents, of which only 260,315 remained after applying this filter.

To further refine the dataset, we removed all entries for which the Portuguese text were duplicated, contained ASCII characters that are not used in the Portuguese language (like ``ø'' or ``ž''), or included repetitive templates that were over-represented in the dataset (steps ``Duplicates'', ``Invalid Chars'' and ``Patterns'' in Figure~\ref{fig:filter_pipeline}). For example, we identified 1,132 documents that began with ``Lista de alterações recentes'' (``List of recent changes'' in English) from a badly scrapped page and 221 documents that started with ``Filtrar por'' (``Filter by''), referring to search engine filters.

Additionally, we filtered out all documents exceeding 900 tokens in the combined source and target texts\footnote{Token count was determined using the \texttt{LLama3} tokenizer.}. This ensures that, regardless of the prompt template used during training, no entry exceeds 1024 tokens. The data loss from this step was minimal, as the dataset primarily consists of single-paragraph documents without line breaks. In fact, documents exceeding 900 tokens accounted for only 0.54\% of the entire dataset. However, removing these documents significantly improved training speed by enabling a larger batch size. When deploying this model, it is crucial to be aware of this limitation and implement strategies to handle larger inputs, such as sentence splitting.

The final step of our filtering pipeline involves a series of miscellaneous checks, such as ensuring that each text contains a matching number of opening and closing brackets and quotation marks. For detailed information about the operations performed, we refer to our code repository. 

We refer to the final parallel corpus as PTradutor. Table~\ref{tab:ptradutor} presents relevant statistics of the dataset after filtering, including the number of tokens and the domains covered. The code to replicate the dataset is available as open-source\footnote{\url{https://github.com/hmosousa/ptradutor}}, and the final corpus is publicly accessible on HuggingFace\footnote{\url{https://huggingface.co/datasets/hugosousa/PTradutor}}.

\begin{table*}
  \centering
  \resizebox{\textwidth}{!}{%
    \begin{tblr}{
      colspec = {Q X[0.5] Q  X[0.5] QQQQ  X[0.5] QQQQ},
      width = \textwidth,
      column{1} = {l},
      column{2-13} = {r},  
      row{1,2} = {c},
      cell{1}{1} = {r=2}{c},
      cell{1}{3} = {r=2}{c},
      cell{1}{5} = {c=4}{c},
      cell{1}{10} = {c=4}{c},
      hline{1,12} = {-}{0.08em},
      hline{3} = {-}{0.05em},
      hline{2} = {5-18}{0.05em},
      hline{10} = {-}{0.05em},
          hline{10} = {2}{-}{},
          hline{11} = {-}{0.08em},
        }
      \textbf{Domain} & & \textbf{\# Docs} & & \textbf{\# Tokens PT} &              &               &                &  & \textbf{\# Tokens EN} &              &              &                \\
                      & &                  & & \textbf{Min}          & \textbf{Max} & \textbf{Mean} & \textbf{Total} &  & \textbf{Min}          & \textbf{Max} & \textbf{Avg} & \textbf{Total} \\
      Journalistic    & & 1,250,982        & & 45                    & 511          & 202.9         & 253,767,361    &  & 25                    & 433          & 150.3        & 188,072,054    \\
      Literature      & & 12,082           & & 51                    & 510          & 121.0         & 1,461,651      &  & 37                    & 360          & 89.8         & 1,085,296      \\
      Web             & & 9,006            & & 44                    & 555          & 224.7         & 2,024,062      &  & 28                    & 416          & 167.1        & 1,504,751      \\
      Politics        & & 477              & & 53                    & 524          & 244.9         & 116,836        &  & 36                    & 380          & 171.5        & 81,801         \\
      Legal           & & 282,870          & & 44                    & 451          & 87.1          & 24,635,676     &  & 25                    & 385          & 64.9         & 18,346,240     \\
      Social Media    & & 163,585          & & 41                    & 129          & 71.0          & 11,622,673     &  & 26                    & 121          & 55.2         & 9,025,327      \\
      DSL-TL (news)   & & 1,734            & & 14                    & 135          & 63.6          & 110,334        &  & 10                    & 108          & 47.2         & 81,821         \\
      \textbf{All}    & & 1,719,002        & & 14                    & 555          & 170.8         & 293,628,259    &  & 10                    & 433          & 126.9        & 218,115,469    \\
    \end{tblr}
  }
  \caption{Statistics describing the PTradutor corpus, including the number of documents (\#~Docs) and the minimum, maximum, average, and total number of tokens per domain.}
  \label{tab:ptradutor}
\end{table*}

\section{Experimental Setup}
\label{sec:exp_setup}
In this section, we describe the training paradigm and model variations used in this study.  
We primarily focus on fine-tuning small-sized LMs, since smaller models are easier for the community to deploy and use. 

\subsection{Models}

To align with the pre-training objective of language models (LMs), we model the translation task as a causal language modeling problem, conditioned on both the translation prompt and the input English text. Specifically, given a translation prompt template \textit{\( \text{prompt} = T(\text{text}_\text{en}) \)} (with \textit{\( \text{text}_\text{en} \)} being the text in English) and a parallel corpus of Portuguese and English, we fine-tune a pre-trained model \textit{\( LM(\textit{prompt}) \)} to generate the translated version of the English text in European Portuguese, \textit{\( \text{text}_\text{pt} \)}. The language models we use are as follows:

\begin{description}
    \item \textbf{Gemma-2} The latest installment in Google's model series, Gemma-2, includes models with 2, 9, and 27 billion parameters~\cite{team2024gemma2}. These models are recognized for their strong performance in reasoning and problem-solving tasks. We use the 2 billion parameter model from this series.
    
    \item \textbf{Phi-3-mini} Developed by Microsoft~\cite{Abdin2024}, the Phi-3 models are compact transformer decoder architectures~\cite{Vaswani2017} with a great percentage of the training corpus being composed of syntactic data. We employ the model with 3.8 billion parameters.
    
    \item \textbf{LLaMA-3} This is the third iteration of Meta's large language model series~\cite{AI2024}. LLaMA-3 models are noted for their enhanced reasoning abilities, logical consistency, and reduced hallucination compared to earlier versions. We use the smallest model from this series, which has 8 billion parameters.
\end{description}

Since the goal is to fine-tune the models to follow a specific instruction (translate the English text to European Portuguese), we use the instruct-tuned version of all models as the starting checkpoint. 
The prompt templates for each model are designed by following the models' guidelines, with the system message of ``\texttt{You are a translator from English to European Portuguese}''  (in case the model supports a system message) and the user prompt ``\texttt{Translate this text from English to European Portuguese: \textit{\( \text{text}_\text{en} \)}}''.

\subsection{Training Approaches}

We utilize two types of instruction fine-tuning:

\textbf{Full fine-tuning} This is similar to the pre-training paradigm, in which given a prompt as instruction, the LM is trained end-to-end, updating all the parameters during optimization. This method although providing better results is often costly, due to the large size of current LMs.

\textbf{Parameter efficient fine-tuning (PEFT)} To decrease the computational and storage costs methods for PEFT~\cite{Xu2023} enable efficient adaptation of LMs by only fine-tuning a small number of extra model parameters instead of all the model's parameters. LoRA~\cite{Hu2022,Xu2024} is one of the most popular methods in this domain, which achieves this by injecting trainable low-rank matrices into each layer of the transformer architecture. In addition to full fine-tuning, we also train the LoRA variant of each model.
\section{Evaluation}
\label{sec:eval}
In this section, we provide a detailed evaluation of our system against various baselines, on two European Portuguese benchmarks. 
The code for training and evaluation, as well as the trained checkpoints of our models, is available in our repository\footnote{\url{https://github.com/hmosousa/tradutor}}.

\subsection{Metrics}

For a comprehensive evaluation, we include both classical and embedding-based metrics.

\paragraph{N-gram based metrics} Classical machine translation metrics~\cite{Popovic2015,Papineni2002,Lin2004,Banerjee2005} usually focus on n-gram overlap between the reference translation and the translation generated by the system. Although by design this metrics fail to recognize semantic similarity beyond the lexical level, they remain widely adopted in academic research. In this work, we incorporate the two most common metrics: BLEU~\cite{Papineni2002} and ROUGE~\cite{Lin2004}.


\paragraph{Learnable metrics}
These methods focus on directly learning human judgment through training. 
For this family, we include COMET~\cite{Rei2020}, which leverages a pre-trained multilingual model. Although COMET can function as a reference-less metric, in this work, we report results exclusively for its direct assessment variant.


\paragraph{Language variety metric}
To assess if the text produced by our translation system is indeed in European Portuguese, we use a Portuguese language variant classifier~\cite{Sousa2025}, which distinguishes between Brazilian and European Portuguese.
After translation of the benchmark data, we employ the classifier to label all generated texts and to compute the percentage of documents that are labeled as European Portuguese. 
Since the classifier might contain intrinsic errors and bias, we also compute the percentage of documents labeled as European Portuguese in the reference translations. 
This step is crucial for handling cases where the translated text for the two variants might be identical. 
In such scenarios, the classifier might incorrectly classify the variety as the wrong one due to the lack of distinguishing features in the text. 
By comparing the results of our system with the reference translations, we can correct for this potential bias and obtain a more accurate assessment of how well our system produces European Portuguese. 
We refer to the ratio of these two percentages as the VID score, which serves as a measure of the system's effectiveness in generating European Portuguese text.

\subsection{Test Benchmarks}
As a low-resource language variant, the number of benchmarks that include European Portuguese is limited. 
In this study, we use two high-quality publicly available datasets that feature this variant:
\begin{itemize}
    \item \textbf{FRMT}: This dataset is specifically designed to contain regional variants of Portuguese and Chinese~\cite{Riley2023}, containing human translations of sentences from English Wikipedia articles that were manually translated to European and Brazilian Portuguese. 
    
    \item \textbf{NTrex}: The dataset consists of high-quality translations by speakers who are bilingual in English and in one of the 128 target languages, including 123 documents and 1,997 sentences for each language~\cite{Federmann2022}.
\end{itemize}

\subsection{Baselines}
We compare our models to three sets of baselines:

\paragraph{Closed Baselines}
We include industry-standard systems for Portuguese translation, including Google Translate and DeepL. Recently, Google Translate introduced a model specifically designed for European Portuguese, referred to as Google$_{pt}$, which we include in our evaluation alongside the original model that does not distinguish between Portuguese varieties, referred to as Google$_{br}$.

\paragraph{Open Baselines}
For open-source systems, we evaluate ArgosTranslate\footnote{\url{https://github.com/argosopentech/argos-translate}}, which uses OpenNMT~\cite{Klein2017} as its backend. Although Portuguese is listed as a supported language, the specific variety is not indicated. We also consider the Opus-MT project~\cite{Tiedemann2020}, another open-source system that provides a model for translating from English to Portuguese. However, like ArgosTranslate, this system does not differentiate between Portuguese varieties.

\paragraph{Zero-shot}
Additionally, we assess the zero-shot capabilities of language models without applying our task-specific fine-tuning to demonstrate the effectiveness of the fine-tuning process.

\subsection{Implementation Details}

All models were trained and evaluated on a server with six A-100 GPUs, each with 40GB of memory. The batch size and training duration varied depending on the memory requirements of each model. In our repository, we provide training and evaluation scripts compatible with the two libraries used to train the language models: \texttt{torchtune}\footnote{\url{https://github.com/pytorch/torchtune/}} and \texttt{transformers}\footnote{\url{https://huggingface.co/}}. While the \texttt{transformers} library was chosen for its practicality, we found it limiting when training larger models. In that scenario, \texttt{torchtune} presented as a reliable alternative with significantly better memory management.
Training runs we executed with early stopping -- using the test set the DSL-TL corpus as validation set -- with patience of 3,000 steps. As a result, the number of training steps varied across models. All LoRA variants were trained with an alpha of 128 and a rank of 256. Detailed training configurations can be found in our repository.

The parameter setup is as follows:

\begin{itemize}
    \item \textbf{Phi-3:} For both the LoRA and full fine-tuning of Phi-3 models, we used a batch size of 512, a learning rate of 2e-5, a weight decay of 0.1, and a warm-up of 1,000 steps.
    \item \textbf{Gemma-2:} For both variants, the learning rate was set to 2e-5 with a weight decay of 0.1. The full fine-tuned model was trained with a 1,000-step warm-up and a batch size of 512, while the LoRA variant had 500 warm-up steps and a batch size of 256.
    \item \textbf{LLaMA-3:} Both variants were trained with a batch size of 256 and a learning rate of 2e-5. The LoRA variant additionally includes a warm-up of 100 steps and a weight decay of 0.1 on the learning rate.
\end{itemize}

\subsection{Results}
\label{sec:results}
\begin{table*}
  \centering
  \resizebox{\textwidth}{!}{%
  \begin{tblr}{
    colspec = {X[3.2] X[3] X[0.05]S[table-format=2.2]S[table-format=2.2]QS[table-format=1.3]X[0.05]S[table-format=2.2]S[table-format=2.2]QS[table-format=1.3]},
    width = \textwidth,
    cell{1}{2} = {r=2}{c},
    cell{1}{4} = {c=4}{c},
    cell{1}{9} = {c=4}{c},
    cell{3}{1} = {r=3}{},
    cell{6}{1} = {r=2}{},
    cell{8}{1} = {r=3}{},
    cell{11}{1} = {r=3}{},
    cell{14}{1} = {r=3}{},
    hline{1,17} = {-}{0.08em},
    hline{3} = {-}{0.05em},
    hline{6,8,11,14} = {-}{dotted},
  }
                              & \textbf{Model}     &    & {{{\textbf{FRMT}}}} &                        &                      &                    &   & {{{\textbf{NTrex}}}} &                        &                      &                    \\
                              &                    & ~  & {{{\textbf{BLEU}}}} & {{{\textbf{ROUGE-L}}}} & {{{\textbf{COMET}}}} & {{{\textbf{VID}}}} & ~ & {{{\textbf{BLEU}}}}  & {{{\textbf{ROUGE-L}}}} & {{{\textbf{COMET}}}} & {{{\textbf{VID}}}} \\
  \textbf{\textbf{Close \newline Baselines}} & Google$_{br}$      &    & 43.20 &  68.43 & 87.44{\scriptsize \(\pm\)0.25} & 0.445 & & 35.80  & 63.44 &  86.88{\scriptsize \(\pm\)0.32} & 0.361  \\
                              & Google$_{pt}$      &    & 47.81 &  71.66 & 87.87{\scriptsize \(\pm\)0.25} & 0.956 & & 39.92  & 66.72 &  87.17{\scriptsize \(\pm\)0.33} & 0.900  \\
                              & DeepL              &    & \underline{49.77} & \underline{72.44} & \underline{88.48}{\scriptsize \(\pm\)0.23} & 0.999 & & \underline{44.76}  & \underline{67.77} & \underline{87.96}{\scriptsize \(\pm\)0.31} & \underline{0.997}  \\
  \textbf{\textbf{Open \newline  Baselines}} & Argos      &    & 38.39  &  65.07 & 83.99{\scriptsize \(\pm\)0.35}  &  0.511  & &  30.44 &  58.72  &  80.30{\scriptsize \(\pm\)0.54}  &  0.446 \\
                                      & Opus-MT    &    & 40.41  &  66.25 & 85.67{\scriptsize \(\pm\)0.31}  &  0.413  & &  32.99 &  60.24  &  83.81{\scriptsize \(\pm\)0.46}  &  0.229 \\
  \textbf{\textbf{Zero-shot}} & Gemma-2            &    & 25.37 &  49.56 & 75.66{\scriptsize \(\pm\)0.51} & 0.807 & & 17.47  & 41.34 &  69.46{\scriptsize \(\pm\)0.66} & 0.858  \\
                              & Phi-3              &    & 17.59 &  43.99 & 57.90{\scriptsize \(\pm\)0.56} & 0.942 & & 13.16  & 38.38 &  54.62{\scriptsize \(\pm\)0.66} & 1.003  \\
                              &  LLaMA-3            &    & 31.47 &  60.61 & 82.95{\scriptsize \(\pm\)0.40} & 0.811 & & 21.09  & 52.34 &  78.70{\scriptsize \(\pm\)0.60} & 0.805  \\
  \textbf{\textbf{LoRA}}      & Gemma-2            &    & 19.83 &  56.87 & 79.62{\scriptsize \(\pm\)0.64} &  0.530  & & 14.41 &  49.42 & 76.18{\scriptsize \(\pm\)0.81} &  0.514 \\
                              & Phi-3              &    & 24.70 &  53.34 & 72.19{\scriptsize \(\pm\)0.58} & \underline{\textbf{1.178}} & & 20.10  & 48.38 &  67.67{\scriptsize \(\pm\)0.73} & \underline{\textbf{1.203}}  \\
                              &  LLaMA-3           &    & 25.42 &  51.51 & 74.06{\scriptsize \(\pm\)0.56} & 1.092 & & 17.74  & 41.24 &  67.96{\scriptsize \(\pm\)0.73} & 1.140  \\
  \textbf{\textbf{FFT}}       & Gemma-2            &    & 33.76 &  66.41 & 85.25{\scriptsize \(\pm\)0.35} & 1.066 & & 18.35  & 59.75 &  82.62{\scriptsize \(\pm\)0.54} & 1.049  \\
                              & Phi-3              &    & 38.16 &  66.31 & 85.35{\scriptsize \(\pm\)0.34} & 1.055 & & 27.89  & 60.18 &  82.91{\scriptsize \(\pm\)0.49} & 1.031  \\
                              &  LLaMA-3            &    & \textbf{41.12} &  \textbf{66.92} & \textbf{86.12}{\scriptsize \(\pm\)0.28} & 0.968 & & \textbf{35.76}  & \textbf{62.02} &  \textbf{84.42}{\scriptsize \(\pm\)0.42} & 0.933  \\           
  \end{tblr}
}
\caption{Model effectiveness was assessed using the FRMT and NTrex benchmarks. Confidence intervals for the COMET metric were computed using a $t$-distribution with a 95\% confidence level. The best results among open-source systems are highlighted in bold, while the best overall results are underlined.}
\label{tab:results}
\end{table*}

The result of our evaluation is shown in Table~\ref{tab:results}, where the best overall values are marked with an underline, and the most effective open-source systems are marked in bold. 
The general trend on both datasets is quite similar and high-performing models maintain a stable performance across both benchmarks. 
Yet, values for NTrex are slightly below FRMT for all systems, indicating a harder benchmark.
In the following, we describe our main findings and highlight avenues for future research.

\paragraph{LoRA models:} These variants effectively learn the vocabulary of European Portuguese, but their overall translation quality remains subpar. 
This discrepancy is evident as they score highly on the VID metric, nevertheless, the BLEU, ROUGE-L, and COMET metrics suggest that they struggle with generating high-quality text. 
Upon closer inspection of the generated translations, we found that both the Phi-3 and LLaMA-3 models, when trained with LoRA, tend to enter a repetition loop, where the same token is generated repeatedly until the process is interrupted. 
This suggests that for a medium size language model translation is a complex task, and simply adding adapter parameters is insufficient to fully capture its nuances. 
This issue, combined with the fact that the early stopping criteria were reached quickly during training (training loss had plateaued while the evaluation was increasing), suggests that the models may require increased capacity (by adjusting the alpha and rank parameters) to better learn from the training data.

\paragraph{Full fine-tuning (FFT):}
The fully fine-tuned models sacrifice their mastery of European Portuguese nuances in favor of high-quality, coherent translations. 
These models yield more moderate scores for the VID metric while achieving significantly higher text quality metrics. 
In particular, the fine-tuned LLaMA-3 model beats all open source software on all metrics and produces results comparable to Google$_{br}$ in terms of BLEU and ROUGE-L on both benchmarks, only falling short on the COMET metric. 

It is important to note that the COMET metric, as a learnable metric, is subject to the same biases that affect any trainable neural model. Since the training data does not differentiate between European and Brazilian Portuguese and most open-source resources are skewed toward Brazilian Portuguese, this bias may have influenced the scores produced by this metric.
The slight difference between Google$_{br}$ and fine-tuned LLaMA-3 models suggests a potential bias in the COMET score toward Brazilian Portuguese.

The overall performance of the fully fine-tuned models has a direct correlation with model size, where our largest model, LLaMA-3, with 8 billion parameters outperforms the smaller models of Phi-3 (3.8 billion parameters) and Gemma-2 (2 billion parameters). 
This behavior is typical for large language models and indicates that increasing their size can lead to better results. This indicates that by applying this methodology to even larger models, it may be possible to achieve results that are on par with industry-standard systems. 

Since our focus is on translation specific to European Portuguese, it is important to examine the VID scores across both benchmarks.
Our best model, LLaMA-3, once again beats all open source baselines and achieves scores significantly higher than Google$_{br}$ and is comparable to Google$_{pt}$ and DeepL. This suggests that the proposed methodology is effective in achieving the targeted goal of producing text specific to a language variety. 
Even our smaller-size models, perform comparable to open-source baselines on translation metrics, dramatically improving on the VID score. 

It is true that in terms of text quality metrics, the LLaMA-3 model still lags behind the European Portuguese-specific industry models, namely Google$_{pt}$ and DeepL. However, it is important to emphasize that our goal was not to beat the specialized model from industry, but to propose a computationally efficient, adaptable, and resource-efficient method for adapting small language models to translate specific language varieties. Surpassing the current open-source software and achieving a score close to industry-level models, which benefit from dedicated teams of experts and annotators for each language, is a significant accomplishment.

\section{Related Work}
\label{sec:related_work}
Despite the availability of industry-level translation systems like Google Translate\footnote{\url{https://translate.google.com/} } and DeepL\footnote{\url{https://deepl.com/} } and a handful of open-source software with unclear language variant definition~\cite{Klein2017,Tiedemann2020},
there are, to the best of our knowledge, no other translation models specifically dedicated to European Portuguese in the literature. 
Like many other industry systems, these models lack transparency, as neither their internal workings nor the data used for training are publicly accessible.

Since this work focuses on Portuguese, we begin by reviewing relevant research in this area. \citet{Lakew2018} explore NMT from English into four pairs of language varieties, including European and Brazilian Portuguese, unlike our focus on Portuguese to English. 
They train a transformer model using transcripts of movies and TED talks but do not open-source their models for comparison. The work primarily highlights the challenges of low-resource settings and the initial steps to address these issues.

Another work in this direction is from \citet{CostaJussa2018}, which aims to train a translation model between standard national varieties of the same language, namely between Brazilian and European Portuguese.
They provide a taxonomy of distinctive characteristics between these two language varieties, which we refer curious readers to for a deeper understanding of the language varieties discussed in this study.

Other previous work in this direction explored a variety-targeted MT system in other languages, which contains varieties or dialects specific to a region.
It is worth noting that although some of these works are similar to textual style transfer in methodology, they are different in the task definition~\cite{Jhamtani2017}.

One of the earlier works in this area focuses on the development of translation systems for Arabic dialects~\cite {Zbib2012,Sajjad2013}. 
Efforts for Arabic took off with the introduction of AraBench benchmark~\cite{Sajjad2020}, an evaluation suite for dialectal Arabic to English MT.

Similar efforts have been undertaken for the low-resource language family of Swiss German, which is widely spoken in Switzerland, but rarely written~\cite{Honnet2018,Scherrer2016}. 
These systems focus on normalizing Swiss German to the standard variant of High German.

Similarly to Arabic and German other languages, such as Chinese, Russian, Hindi and Turkish, are also slowly finding their way into this study~\cite{Wan2020,Kumar2021,Nguyen2017,Durrani2010}.

Another area of research loosely related to ours involves adapting language models to low-resource language varieties. 
As previously mentioned, the effectiveness of large language models in this domain is often limited by the scarcity of representative datasets~\cite{Alam2024}. 
Recent efforts have focused on creating specialized language models for specific dialects~\cite{Ondrejova2024,Faisal2024,Nguyen2024,Lopes2024}, mainly through data augmentation or the introduction of new datasets. 
Although these are not translation systems and therefore are not directly comparable to our work, our methodology and provided resources can also be used to provide artificial training data. 

\section{Conclusion \& Future Work}
\label{sec:conculsion}
In this paper, we present a methodology for creating a parallel corpus and training a translation model tailored to a low-resource language variety. Specifically, we developed and open-sourced the largest European Portuguese-English parallel corpus, along with European Portuguese-specific translation models. Our extensive evaluation demonstrates the effectiveness of our approach and the fidelity of the generated translations to the desired language variety. Thanks to the proposed methodology we managed to achieve performance on par with industry-level translation systems with minimal resources and limited computation.

In future work, it would be interesting to investigate the impact of different generation configurations on the translations produced by our model. In this study, we used greedy decoding, but other generation techniques, such as beam search, could yield better results. Another promising direction is to explore prompt optimization both before and after model training. This approach has been shown to improve outcomes in other studies~\cite{Soylu2024} and might enhance our system's performance as well. Finally, we plan to conduct a human evaluation with linguists to identify areas where our model falls short compared to other systems.

\section*{Acknowledgements}

This research is supported by national funding from the Portuguese Foundation for Science and Technology (FCT) under the project with DOI \texttt{\small 10.54499/LA/P/0063/2020}. The authors also acknowledge the support of the StorySense project (DOI \texttt{\small 10.54499/2022.09312.PTDC}) and the advanced computing project PTicola (ID \texttt{\small CPCA-IAC/AV/594794/2023}). Hugo Sousa further acknowledges FCT for funding his PhD grant (ID \texttt{\small 2022.14691.BD}).

\bibliography{references}

\end{document}